\documentclass[10pt,twocolumn,letterpaper]{article}

\usepackage[pagenumbers]{cvpr} 

\usepackage{graphicx}
\usepackage{times}
\usepackage{helvet}
\usepackage{courier}
\usepackage{amsmath}
\usepackage{algorithm}
\usepackage{algorithmic}
\usepackage{csquotes} 
\usepackage{color}
\usepackage{paralist}
\usepackage{amssymb}
\usepackage{indentfirst}
\usepackage{pifont}
\usepackage{float}
\usepackage{multirow}
\usepackage{mathrsfs}
\usepackage{makecell}

\usepackage{mathrsfs}
\usepackage{textcomp,booktabs}
\usepackage{amssymb}
\usepackage{pifont}
\usepackage[misc]{ifsym}
\newcommand{\cmark}{\ding{51}}%
\usepackage{soul}
\usepackage{colortbl}
\usepackage[dvipsnames]{xcolor}
\usepackage{mathrsfs}
\usepackage{color}
\usepackage{xcolor}
\definecolor{citecolor}{HTML}{0071bc} 
\definecolor{SeaGreen4}{RGB}{0,205,102} 
\definecolor{SlateBlue}{RGB}{106,90,205} 
\definecolor{DarkRed}{RGB}{178,34,34} 
\usepackage[colorlinks, linkcolor=red,  anchorcolor=blue, citecolor=citecolor]{hyperref}

\usepackage{colortbl}
\definecolor{mygray}{gray}{.9}
\definecolor{mypink}{rgb}{.99,.91,.95}
\definecolor{mycyan}{cmyk}{.3,0,0,0}

\usepackage{color}
\usepackage{xcolor}
\definecolor{citecolor}{HTML}{0071bc} 
\definecolor{SeaGreen4}{RGB}{0,205,102} 
\definecolor{SlateBlue}{RGB}{106,90,205} 
\definecolor{DarkRed}{RGB}{178,34,34}

\usepackage[capitalize]{cleveref}
\crefname{section}{Sec.}{Secs.}
\Crefname{section}{Section}{Sections}
\Crefname{table}{Table}{Tables}
\crefname{table}{Tab.}{Tabs.}

\def\paperID{*****} 
\def\confName{CVPR}
\def\confYear{2025}

\title{ Towards Low-Latency Event Stream-based Visual Object Tracking: A Slow-Fast Approach }

\author{Shiao Wang$^{1}$, Xiao Wang$^{1}$\thanks{Corresponding Author: Xiao Wang \& Bo Jiang}, Liye Jin$^{1}$, Bo Jiang*$^{1}$, Lin Zhu$^{2}$, Lan Chen$^{3}$, Yonghong Tian$^{4,5,6}$, Bin Luo$^{1}$ \\ 
${^1}${School of Computer Science and Technology, Anhui University, Hefei, China} \\
${^2}${Beijing Institute of Technology, Beijing, China} \\
${^3}${School of Electronic and Information Engineering, Anhui University, Hefei, China} \\
${^4}${Peng Cheng Laboratory, Shenzhen, China} \\ 
${^5}${School of Computer Science, Peking University, China} \\ 
${^6}${School of Electronic and Computer Engineering, Shenzhen Graduate School, Peking University, China} \\ 
\textit{e24101001@stu.ahu.edu.cn}, \textit{\{xiaowang, chenlan, jiangbo, luobin\}@ahu.edu.cn}, \\ 
\textit{jinliye@stu.ahu.edu.cn}, \textit{\{linzhu, yhtian\}@pku.edu.cn}
}

\begin{document}
\maketitle

\begin{abstract}
Existing tracking algorithms typically rely on low-frame-rate RGB cameras coupled with computationally intensive deep neural network architectures to achieve effective tracking. However, such frame-based methods inherently face challenges in achieving low-latency performance and often fail in resource-constrained environments. Visual object tracking using bio-inspired event cameras has emerged as a promising research direction in recent years, offering distinct advantages for low-latency applications. 
In this paper, we propose a novel Slow-Fast Tracking paradigm that flexibly adapts to different operational requirements, termed SFTrack. The proposed framework supports two complementary modes, i.e., a high-precision slow tracker for scenarios with sufficient computational resources, and an efficient fast tracker tailored for latency-aware, resource-constrained environments.
Specifically, our framework first performs graph-based representation learning from high-temporal-resolution event streams, and then integrates the learned graph-structured information into two FlashAttention-based vision backbones, yielding the slow and fast trackers, respectively. The fast tracker achieves low latency through a lightweight network design and by producing multiple bounding box outputs in a single forward pass. 
Finally, we seamlessly combine both trackers via supervised fine-tuning and further enhance the fast tracker’s performance through a knowledge distillation strategy.
Extensive experiments on public benchmarks, including FE240, COESOT, and EventVOT, demonstrate the effectiveness and efficiency of our proposed method across different real-world scenarios.
The source code has been released on \url{https://github.com/Event-AHU/SlowFast_Event_Track}. 
\end{abstract}

\section{Introduction} 

Visual Object Tracking (VOT) aims to continuously localize a target in subsequent frames, given its initial bounding box in the first frame. 
Most existing trackers~\cite{chen2021transt, wang2021TNL2K} rely on mature RGB camera technology and have achieved strong performance across applications such as autonomous driving, surveillance, and drone photography. However, in real-world deployments, RGB cameras are vulnerable to environmental challenges like overexposure, low light, motion blur, and fast motion, often resulting in tracking failures and limiting the robustness of RGB-only systems. 

\begin{figure*}
\center
\includegraphics[width=0.95\linewidth]{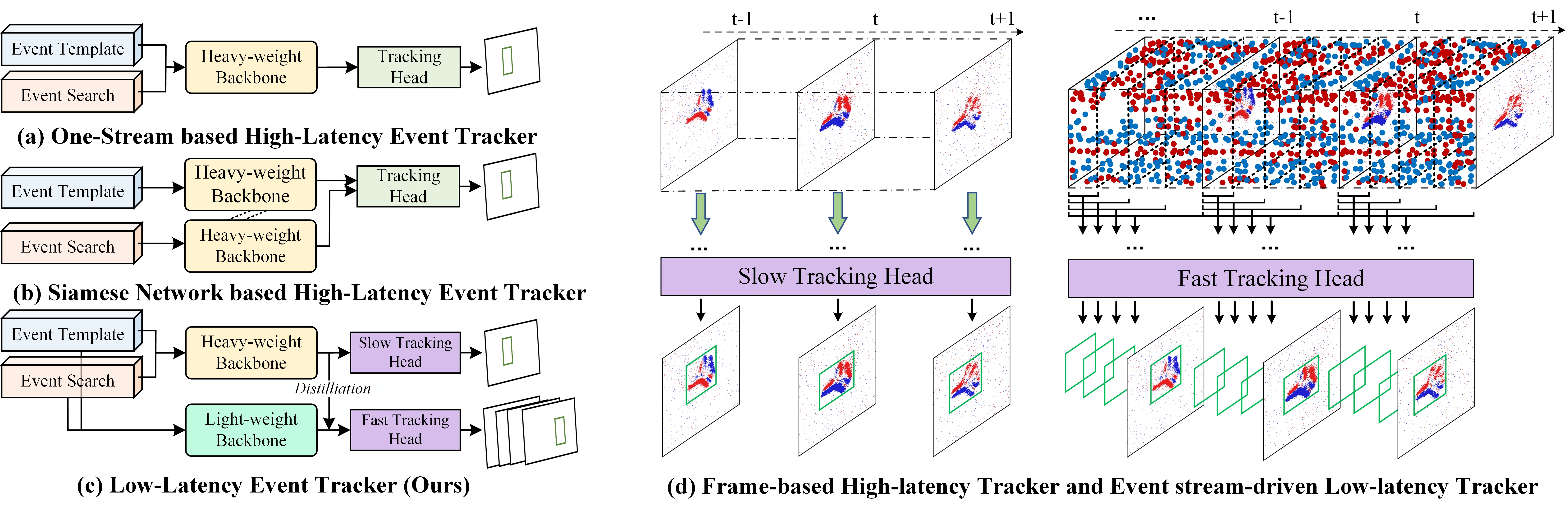}
\caption{Existing event-based tracking methods either adopt (a) \textit{a one-stream based tracking} or (b) \textit{a Siamese network based tracking}, both of which suffer from high latency. We propose (c) \textit{a dual-head tracking framework} designed to achieve \textbf{low-latency tracking} in line with real-world requirements. (d) A comparison between the conventional frame-based high-latency tracker and the event stream-driven low-latency tracker.} 
\label{firstIMG}
\end{figure*} 

Bio-inspired event cameras offer an alternative paradigm by asynchronously capturing pixel-level intensity changes, enabling high temporal resolution, high dynamic range, and inherent privacy protection. These advantages make them strong competitors to conventional RGB cameras, especially under extreme conditions~\cite{gallego2020event}. Motivated by this, an increasing number of researchers have explored the integration of event streams with RGB images to mitigate performance degradation in challenging environments. For instance, Zhang et al. propose AFNet~\cite{zhang2023frame} to combine the RGB and event data via multimodal alignment and fusion modules. Wang et al.~\cite{wang2023visevent} introduce a simple yet effective RGB-Event tracking algorithm using a cross-modality fusion Transformer network. Despite these advancements, the fusion of multimodal data, especially in architectures dominated by Vision Transformers (ViT)~\cite{dosovitskiy2020image} and large foundation models~\cite{wang2023MMPTMSURVEY}, often incurs substantial computational overhead. This significantly hampers real-time or low-latency performance in visual object tracking tasks, limiting the practicality of such systems in time-sensitive applications.


To achieve efficient tracking, researchers have increasingly turned to event-based unimodal tracking methods to mitigate the computational overhead associated with multimodal data processing. As illustrated in Fig.~\ref{firstIMG} (a) and (b), influenced by current deep learning networks (i.e., Convolutional Neural Network (CNN)~\cite{lecun1998gradient}, ViT~\cite{dosovitskiy2020image}, 
and Mamba~\cite{gu2023mamba}), existing event-based trackers~\cite{chen2019asynchronous, wang2024event, zhang2022STN} typically adopt either one-stream tracking frameworks or conventional Siamese network architectures to achieve effective object tracking. Although both tracking paradigms perform well, they are inherently limited by high latency and frame-level tracking, which prevents them from fully leveraging the high temporal resolution that event cameras offer, thereby constraining their real-world applicability. Therefore, \emph{designing a tracker capable of meeting different real-world requirements, including both high precision and low latency, has become an urgent research priority.}

\begin{figure}
\center
\includegraphics[width=0.48\textwidth]{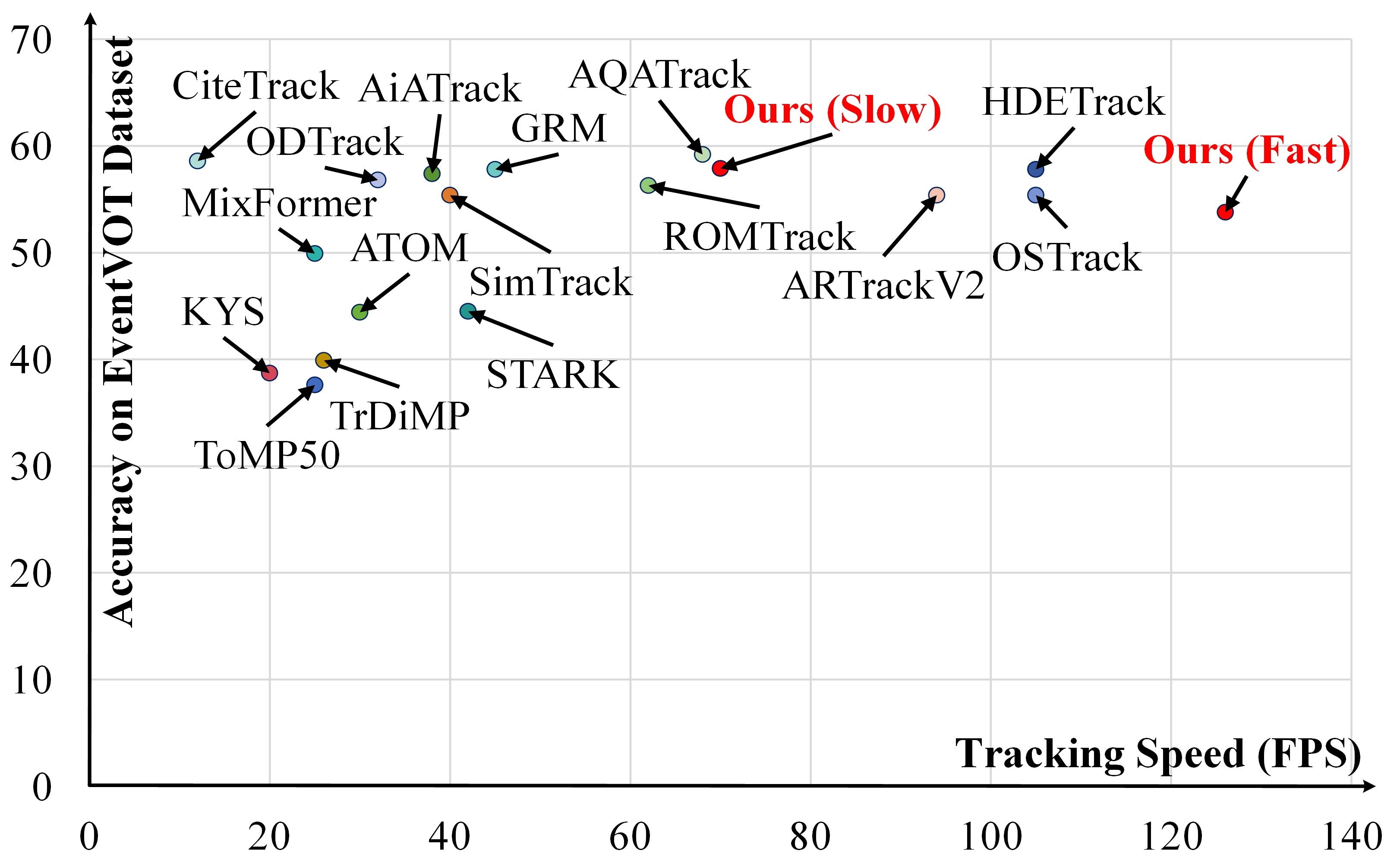}
\caption{Comparison of Accuracy and Tracking Speed with other Trackers.}  
\label{PRandFPS}
\end{figure}

In this work, we introduce a novel Slow-Fast Tracking framework, termed SFTrack, as shown in Fig.~\ref{firstIMG} (c), which seamlessly combines high-precision and low-latency tracking paradigms. 
The slow tracker benefits from multi-scale feature fusion, combining high-temporal-resolution event streams and event frames for improved accuracy, while the fast tracker achieves low latency by extracting features from event stream segments of varying sparsity to produce multiple tracking results. 
Our framework adopts a two-stage training strategy as shown in our framework. 
In the first stage, we independently train two specialized trackers, i.e., a high-precision slow tracker and a lightweight fast tracker optimized for low-latency scenarios.
Specifically, we first project the cropped template and search region from the input event frames into visual token representations via a projection layer. In parallel, we construct a spatiotemporal graph from event points within a specific temporal window to capture fine-grained motion cues with high temporal fidelity. These graphs are then fed into Graph Neural Networks (GNNs) to extract multi-scale event features, which are then fused with visual tokens in each tracker, allowing for to benefit from the complementary strengths of visual tokens and event-driven cues. Finally, two dedicated tracking heads are employed to predict the target’s location within the search region. 
In the second stage, we perform a simple yet effective fine-tuning process, which not only unifies the two trackers into a cohesive architecture but also further improves the robustness of the fast tracker through a knowledge distillation strategy, allowing it to approach the accuracy of the slow tracker without sacrificing tracking speed. 
Fig.~\ref{firstIMG}(d) presents a comparison between the conventional frame-based high-latency slow tracker and the event stream-driven low-latency fast tracker. The former follows a frame-level tracking paradigm, yielding a single tracking result per frame, whereas the latter exploits the high temporal resolution of event streams to generate multiple tracking results sequentially within a single time window. In Fig.~\ref{PRandFPS}, we compare the accuracy and tracking speed of our method with other State-Of-The-Art (SOTA) approaches, highlighting the robust performance of the slow tracker and the low-latency advantage of the fast tracker.

To sum up, the contributions of this work can be summarized as follows:
\textit{1).} We propose a novel \textbf{Slow-Fast Tracking} (SFTrack) paradigm for event-based visual object tracking. To the best of our knowledge, it is the first dual-head tracker specifically designed to meet high-precision and low-latency requirements for different real-world applications. 
\textit{2).} We fully leverage the high temporal resolution of event cameras by integrating event streams into the tracking process, breaking the limitations of frame-based object tracking and achieving millisecond-level low-latency tracking.
\textit{3).} Extensive experiments on three public benchmark datasets, i.e., FE240hz, COESOT, and EventVOT, fully demonstrate the efficiency and effectiveness of our proposed tracker.

\section{Related Works} 
In this section, we review the most related research topics to our paper, including Event Stream based Tracking, Slow-Fast Learning, and Low-Latency Design. More related works can be found in the following surveys~\cite{gallego2020event, marvasti2021trackSurvey} and paper list \footnote{\url{github.com/wangxiao5791509/Single_Object_Tracking_Paper_List}}.

\subsection{Event Stream based Tracking}  
Event stream-based tracking has emerged as a promising research direction, garnering increasing attention in recent years. 
The early event-based tracker ESVM~\cite{8368143} was proposed by Huang et al. for high-speed moving object tracking. 
After that, Chen et al.~\cite{Chen_2019} introduced ATSLTD, an adaptive event-to-frame conversion algorithm designed for asynchronous tracking. 
EKLT~\cite{gehrig2020eklt} combined frame and event streams for high-temporal-resolution feature tracking. 
In recent years, deep learning architectures have shown their advantages in tracking methods~\cite{wang2023visevent, ye2022Ostrack, tang2022coesot, zheng2024odtrack, huang2024mambafetrack}. 
STNet~\cite{zhang2022spiking} integrated a Transformer and an SNN for spatiotemporal modeling. 
Tang et al.~\cite{tang2022coesot} proposed unified backbones for joint feature extraction and fusion. 
Zhu et al.~\cite{zhu2022learning} improved tracking accuracy and speed through key event sampling and graph network embedding. 
Wang et al.~\cite{wang2024event} proposed a novel hierarchical cross-modality knowledge distillation approach. Different from the above methods, we fully exploit the high temporal resolution characteristics of event stream data by constructing event-based graphs to enhance the temporal feature representation capability.

\subsection{Slow-Fast Learning} 
The Slow-Fast paradigm has emerged as a powerful framework for processing spatiotemporal data. 
The origins of Slow-Fast learning can be traced back to Feichtenhofer et al.~\cite{feichtenhofer2019slowfast}, who proposed the Slow-Fast framework to effectively model both spatial semantics and temporal dynamics in video understanding through dual-pathway processing. 
Subsequently, X3D~\cite{feichtenhofer2020x3d} extended the Slow-Fast concept by proposing a scalable 3D CNN architecture. 
TimeSformer~\cite{bertasius2021space} integrated Slow-Fast with Transformer architectures and proposed a purely attention-based approach for spatiotemporal modeling. 
More recent efforts~\cite{slowfastvgen, shi2025slow} focus on developing models that can learn from multiple modalities. Specifically, AVSlowFast proposed by Xiao et al.~\cite{xiao2020audiovisual} to model vision and sound in a unified representation. 
Tong et al.~\cite{tong2022videomae} proposed VideoMAE, which incorporated the Slow-Fast frame sampling strategy into a self-supervised video pre-training framework.
Shi et al.~\cite{shi2025slow} proposed a Slow-Fast architecture for video multi-modal large language models (MLLMs). 
In this work, we propose a unified Slow-Fast tracking framework for event-based object tracking, adaptable to both resource-rich and resource-constrained scenarios.

\subsection{Low-Latency Design}  
Recent advances in low-latency design have enabled visual trackers to achieve both efficient deployment and competitive accuracy.
For example, Danelljan et al. proposed ATOM~\cite{danelljan2019atom}, which is designed as a lightweight structure to reduce computational complexity. 
LightTrack~\cite{yan2021lighttrack} is proposed to discover lightweight backbone and head networks. 
Following OSTrack~\cite {blatter2023efficient}, research has increasingly focused on efficient, low-latency tracking within the one-stream architecture. For instance, MixFormerV2~\cite{cui2024mixformerv2} lightened a one-stream network through dense-to-sparse and deep-to-shallow distillation. 
Lin et al.~\cite{lin2024tracking} proposed LoRAT, which incorporated parameter-efficient fine-tuning techniques into tracking tasks.
However, the aforementioned methods are all frame-level tracking approaches. In this work, we leverage high-temporal-resolution event streams to enable low-latency tracking.

\begin{figure*}
\center
\includegraphics[width=\linewidth]{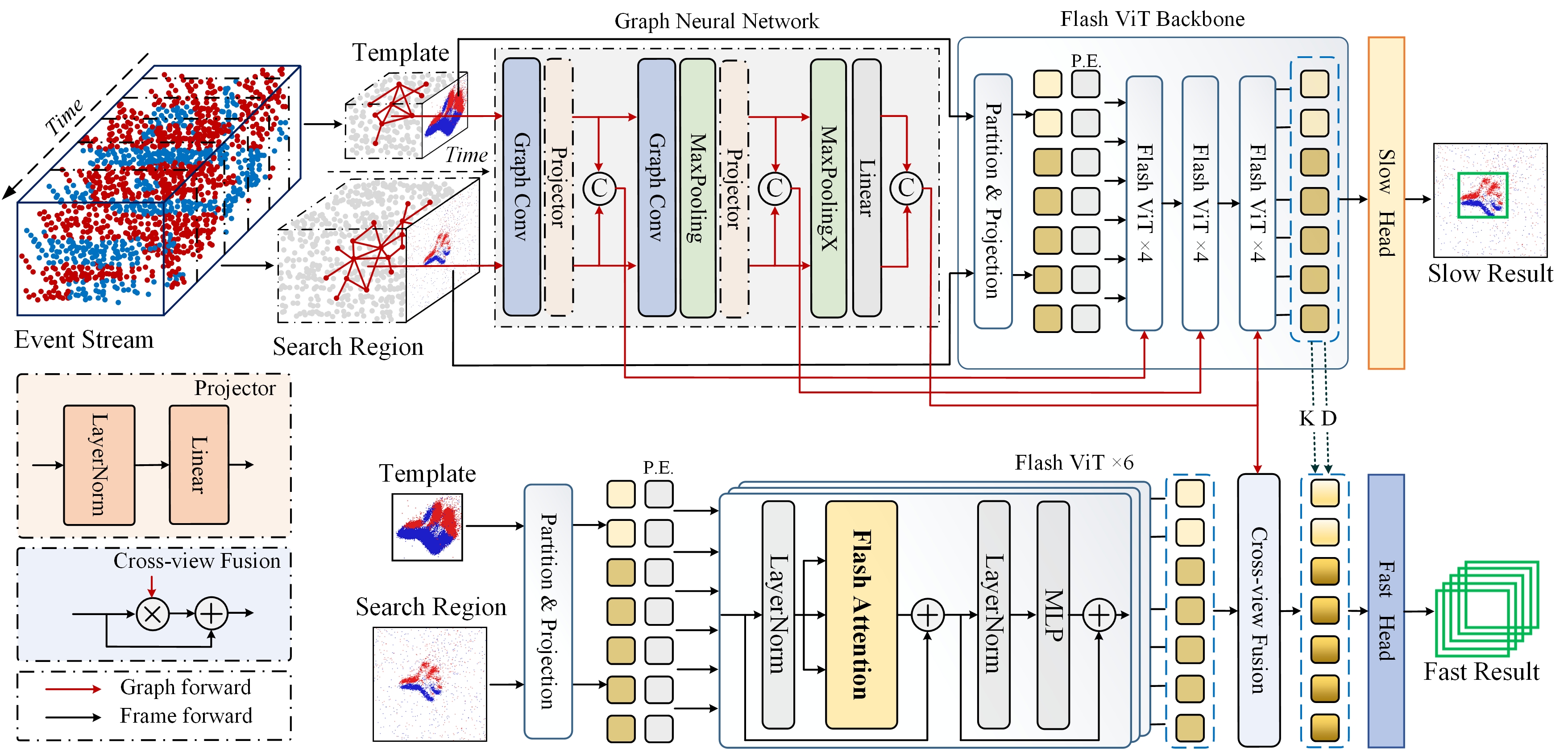}
\caption{\textbf{An overview of our proposed Slow-Fast Framework for Event-based Tracking.} It contains a slow tracker with a 12-layer flash-attention based ViT and a fast tracker with a 6-layer. In addition to the input event frames, both trackers are also fed with event points to optimize their tracking performance. Specifically, we first train the slow and fast tracker separately, enhancing their tracking performance through effective fusion with the event-based graph feature representation. Subsequently, by employing supervised fine-tuning and a knowledge distillation strategy, we not only achieve seamless integration between the two trackers but also significantly improve the performance of the fast tracker. Our Slow-Fast approach demonstrates flexible adaptability to real-world scenarios, whether under resource-rich or resource-constrained conditions.
} 
\label{framework}
\end{figure*}

\section{Methodology} 

\subsection{Overview} 
To accommodate different real-world application requirements, we propose a novel Slow-Fast tracking framework that delivers superior performance in resource-rich scenarios while maintaining low-latency tracking under resource-constrained conditions. The overall process is divided into two sequential stages, as shown in Fig.~\ref{framework}. 
In the \textit{first stage}, we enhance the robustness of the slow tracker by leveraging the high temporal resolution of event stream data through the integration of raw event points. This approach leverages Graph Neural Network (GNN) for feature extraction, while combining with event frame features obtained through Vision Transformer (ViT) for multi-scale feature fusion. Simultaneously,  the event point features are also incorporated into the fast tracker through an effective cross-view fusion mechanism, which enhances feature representation while preserving low-latency tracking. 
In the \textit{second stage}, after obtaining both the slow and fast trackers, we integrate them into a unified framework through supervised fine-tuning, and further enhance the performance of the fast tracker using a knowledge distillation strategy. Detailed descriptions of the network architecture and the Slow-Fast tracking strategy are provided in the following subsections.

\subsection{Input Representation}
Given the event stream $\mathcal{E_P} = \{e_1, e_2, ..., e_M\}$ during the time interval $T$, which contain $M$ discrete event points asynchronously launched where each event point $e_i$ can be represented as quadruple $\{x_i, y_i, t_i, p_i\}$. The $(x_i, y_i)$ is the spatial coordinate of $e_i$, while the $t_i \in [0, T]$ and $p_i \in \{-1,1\}$ denote the timestamp and polarity, respectively. To facilitate deep learning model learning, we stack the event stream into event frames at fixed time intervals $\Delta t$, and obtain the event frames $\mathcal{E} \in \mathbb{R}^{N \times C \times H \times W}$. Here, the $N = T/ \Delta t$ is the number of event frames, the $C$ is the channel dimension, and $H$, $W$ denote the height and width of each event frame.

\subsection{Network Architecture}
In this paper, we propose a novel Slow-Fast tracking framework for event-based visual object tracking. As shown in Fig.~\ref{framework}, a slow tracker and a fast tracker are seamlessly integrated to accommodate diverse real-world requirements.

\noindent $\bullet$ \textbf{Slow Tracker Design.~} 
In real-world scenarios with abundant computational resources, the focus often shifts toward achieving higher precision for more accurate tracking. To achieve this, our slow tracker leverages the dense temporal information inherent in event streams, significantly enhancing tracking accuracy. 

We begin by cropping the template and the search regions from the event frames using the target bounding boxes. These cropped event frames are split into $N=HW/P^2$ patches, where P is the patch size, and then passed through a projection layer to obtain patch embeddings. The embedded tokens from the template and search regions are combined with positional encodings to obtain $Z_0 \in \mathbb{R}^{B \times N_z \times C}$ and $S_i \in \mathbb{R}^{B \times N_s \times C}$, where $B$ is the batch size, $C$ is the channel dimension, $N_z$ and $N_s$ denote the numbers of template tokens and search tokens, respectively. We then concatenate $Z_0$ and $S_i$ to obtain X as the input of the backbone network. Following the OSTrack~\cite{ye2022Ostrack}, a 12-layer ViT with uniform width is employed as the backvision backbone to effectively extract visual features from the event frames. It is worth noting that the standard self-attention mechanism in ViT is replaced with FlashAttention2~\cite{dao2023flashattention2} in our design to enable more efficient feature modeling. The mathematical formula in Flash ViT can be expressed as follows (Layer-Normalization operation is omitted in the equation):
\begin{equation}
\begin{aligned}
    X'_l &= \text{FA}(X_{l-1}) + X_{l-1}, \quad (l = 1, 2, \dots, L) \\
    X_l  &= \text{FFN}(X'_l) + X'_l \\
         &= \text{FFN}\left(\text{FA}(X_{l-1}) + X_{l-1}\right) + \text{FA}(X_{l-1}) + X_{l-1},
\end{aligned}
\end{equation}
where the FA and FFN denote the FlashAttention and feed-forward network, respectively. L is the total number of Flash ViT blocks.
 
To fully leverage the high temporal resolution inherent in event stream data, we also incorporate event points from the time intervals corresponding to each event frame as inputs to the network. This approach aims to enhance the network’s ability to capture fine-grained temporal details and improve overall tracking performance, ensuring that the tracker benefits from both the spatial and temporal information richness of event data. Specifically, we first perform uniform downsampling on the dense event points to obtain $V$, thereby reducing the computational burden of the network. Then, using the k-nearest neighbors (K-NN) graph construction method, we create the graph structure based on the adjacency matrix $A_{ij}$. Subsequently, we feed the graph structure, which contains the sparse event points, into the simple and lightweight 2-layer Graph Convolutional Neural Network (GNN) to extract and model the features, ultimately obtaining the graph-based features of the event points. This process can be represented by the following formula,
\begin{equation}
\begin{aligned}
    F_g &= \text{GCN}(G), \quad G = \{V, E\}, \\
    A_{ij} &= 
    \begin{cases} 
        1, & \text{if } V_j \in \mathcal{N}_k(V_i) \text{ or } V_i \in \mathcal{N}_k(V_j) \\ 
        0, & \text{otherwise}
    \end{cases}
\end{aligned}
\end{equation}
where the $V_i$ and $V_j$ denotes the $i$-th and $j$-th event point in the downsampled event points. $\mathcal{N}_k(.)$ is the K-NN function. $A_{ij}$ is the adjacency matrix, indicating whether there is an edge connection between node $i$ and node $j$. $G$ represents the graph structure, which consists of vertices $V$ and edges $E$. $F_g$ denotes the graph features extracted by GCN layers. Notably, for sufficient information fusion, we extract hierarchical features from multiple GCN layers to capture multi-scale spatiotemporal information within the event graph. To ensure dimensional consistency with the visual features from the Flash ViT backbone, lightweight projection layers are employed to align the multi-scale graph representations with the target embedding space. This process can be mathematically formulated as:

\begin{equation}
\begin{aligned}
    G' &= \text{GCN}_1(G),  \quad F_g^1 = \text{Projector}_1(G'), \\
    G'' &= \text{MaxPooling}(\text{GCN}_2(G')), \quad F_g^2 = \text{Projector}_2(G''), \\
    F_g^3 &= \text{Linear}(\text{MaxPooling}(G'')),
\end{aligned}
\end{equation}
where the $\text{GCN}_i$, $\text{Projector}_i$, and $\text{F}_g^i$ represent the $i$-th layer of GCN, $i$-th Projector and $i$-th graph output feature, respectively. $G'$ and $G''$ are the outputs from the GCNs. The MaxPooling operation transforms the whole graph to a coarser graph $G_c$.

In this work, we refer to AEGNN~\cite{Schaefer22cvpr}, which uses the voxel-grid-based max-pooling~\cite{simonovsky2017dynamic} due to its computational efficiency and simplicity. The method proposed in~\cite{simonovsky2017dynamic} clusters the graph’s vertices by mapping them to a uniformly spaced spatio-temporal voxel grid, where all vertices within the same voxel are grouped into one cluster $C_k$ with cluster centers $k \in V_c$, which from a subset of $V$. Similarly, we employ a voxel grid with dimensions of 12×16×16, where a single node is selected within each voxel and the original edges are reconnected to construct a coarse-grained graph structure. Unlike MaxPooling, MaxPoolingX exclusively performs max-pooling on node features. Following the given cluster assignment, new node features $x_k$ are generated by taking the maximum feature values within each cluster. The mathematical formula is as follows $x_k = \max_{i \in \mathcal{C}_k} x_i$, where the $x_i$ denote the node features attached to each node in $V$.  

The resulting multi-scale graph features are then concatenated with the corresponding shallow, intermediate, and deep features from the Flash ViT backbone, facilitating comprehensive fusion across different levels. After obtaining the multi-scale fused features, we follow the design of OSTrack by feeding the features corresponding to the search region into a center-based tracking head. The target localization is then achieved by the predicted response map.

\noindent $\bullet$ \textbf{Fast Tracker Design.~}
On the other hand, when faced with limited resources, it is challenging to deploy high-complexity models for effective tracking. As a result, the focus often shifts toward achieving low-latency object tracking using a lightweight model, even at the cost of a slight decrease in accuracy. To this end, we design a fast tracker tailored for real-world low-latency scenarios.

Similar to the slow tracker introduced above, we also use both event frames and event point data as the combined input to the fast tracker. To achieve low-latency object tracking, we propose a faster and more lightweight tracking framework with the following three key designs:

\emph{Firstly}, we reduce the computational complexity of the Flash ViT backbone through layer pruning. Specifically, we retain only half of the Transformer layers, significantly decreasing model depth and inference time while preserving sufficient representation capacity for efficient tracking. This pruning strategy ensures high-speed inference, which is essential for latency-aware applications.

\emph{Secondly}, to avoid introducing additional parameters and further reduce computational burden, we design a simple yet effective cross-view feature fusion mechanism. Instead of relying on complex fusion networks, we perform element-wise multiplication to extract shared representations from both event frame and event point features, which are then added to the visual features via a residual connection. This operation enables the network to capture the temporal information embedded in event streams without increasing model size. The cross-view fusion process is formulated as follows,
\begin{equation}
\begin{aligned}
\label{fusion}
{F'} = {F}_\text{v} * {F}_\text{g} + {F}_\text{v},
\end{aligned}
\end{equation}
where ${F}_\text{v}$ and ${F}_\text{g}$ denote the vision features and graph features, respectively.

\emph{Thirdly}, our fast tracker introduces an event graph accumulation strategy for processing event streams. As the event density increases, it dynamically generates graph-structured representations, capturing evolving event patterns over time, yielding a progressive sequence of graph features (from sparse to dense) that inherently reflect the temporal evolution of event data. These temporally structured graph features are sequentially fused with visual features and fed into the tracking head (similar to the slow head) to generate multiple tracking predictions across different time scales. Compared with traditional frame-based trackers, our method can generate multiple desired tracking results from the event stream segment corresponding to a video frame, achieving millisecond-level tracking speed, significantly improving tracking speed, and enabling true low-latency tracking.

These designs effectively balance tracking accuracy and computational efficiency, addressing the critical trade-off commonly faced in real-time visual tracking. By significantly reducing model complexity while preserving essential representational capacity, our framework delivers fast and reliable performance under the constrained computational budgets typical of edge platforms. This makes it especially well-suited for real-world deployment scenarios such as mobile robotics, autonomous driving, and wearable devices, where both low latency and lightweight operation are crucial.

\noindent $\bullet$ \textbf{Unified Supervised Fine-tuning.~}
Through the aforementioned first-stage training, we are able to obtain a high-precision slow tracker and a low-latency fast tracker, each tailored to different performance requirements. The slow tracker focuses on precise localization through a more expressive architecture, while the fast tracker prioritizes rapid inference, enabling efficient tracking in real-time scenarios. Next, we integrate the two trackers through supervised fine-tuning in a unified framework. Specifically, to avoid mutual interference between network parameters, we first freeze the network of the slow tracker and adopt a lower learning rate to fine-tune the fast tracker. In addition, we employ a teacher-student training paradigm, where the slow tracker serves as the teacher network and the fast tracker as the student. The robust features extracted by the teacher are used to guide the learning of the student, enabling effective knowledge transfer. This strategy not only allows for a seamless integration of the two trackers but also enhances the tracking performance of the fast tracker by leveraging the strengths of the slow tracker in a supervisory role. The process of knowledge transfer can be expressed by the following formula,
\begin{equation}
\label{mse}
\mathcal{L}_{\text{KD}} = \frac{1}{n} \sum_{i=1}^{n} ({F}_{f} - {F}_{s})^2
\end{equation}
where the ${F}_{f}$ and ${F}_{s}$ denote the features from the fast tracker and the slow tracker, respectively. $n$ is the number of input samples.

\subsection{Loss Function} 
In the first stage of training, both our slow tracker and fast tracker adopt the same tracking loss as used in OSTrack~\cite{ye2022Ostrack} (i.e., focal loss $L_{focal}$, $L_1$ loss, and GIoU loss $L_{GIoU}$ ). During the second stage of fine-tuning, we further incorporate a knowledge distillation loss to guide the training of the fast tracker. The total loss can be formulated as,
\begin{equation}
\mathcal{L}_{\text{total}} = 
        \lambda_1 \mathcal{L}_{\text{focal}} + \lambda_2 \mathcal{L}_1 + \lambda_3 \mathcal{L}_{\text{GIoU}} + \lambda_4 \mathcal{L}_{\text{KD}},
\end{equation}
where $\lambda_i$ ($i \in [1,4]$) are the weighting coefficients that balance the contributions of the $\mathcal{L}_{\text{focal}}$, $\mathcal{L}_1$, $\mathcal{L}_{\text{GIoU}}$, and $\mathcal{L}_{\text{KD}}$, respectively.

\section{Experiments} 

\subsection{Datasets and Evaluation Metric}  
In this section, we compare with other SOTA trackers on existing event-based tracking datasets, including \textbf{FE240hz}~\cite{zhang2021fe108}, \textbf{COESOT}~\cite{tang2022coesot}, and \textbf{EventVOT}~\cite{wang2024event} dataset. Note that, for the \textbf{FE240hz}~\cite{zhang2021fe108} and \textbf{COESOT}~\cite{tang2022coesot} dataset, we conduct comparisons with other methods by exclusively using unimodal event data as input, and by retraining and testing the competing methods to ensure fairness in the evaluation. A brief introduction to the event-based tracking datasets used in this work is provided below.

$\bullet$ \textbf{FE240hz dataset}: It was collected using a grayscale DVS 346 event camera and consists of 71 training videos and 25 testing videos. The dataset provides over 1.13 million annotations across more than 143K images and their corresponding event data. It also considers various challenging conditions for tracking, such as motion blur and high dynamic range scenarios.

$\bullet$ \textbf{COESOT dataset}: It is a large-scale, category-wide RGB-Event-based tracking dataset that covers 90 object categories and includes 1,354 video sequences, comprising a total of 478,721 event frames. To systematically evaluate tracking robustness, 17 challenging factors are formally defined within the dataset. The dataset is split into 827 videos for training and 527 videos for testing, providing a comprehensive benchmark for advancing event-based tracking research.

$\bullet$ \textbf{EventVOT dataset}: It is the first large-scale, high-resolution event-based tracking dataset captured by a Prophesee camera. The dataset consists of 841 training videos and 282 testing videos, covering 19 categories such as UAVs, basketball, and pedestrians, and so on. Each video is uniformly divided into 499 frames to facilitate professional annotation. A total of 14 challenging attributes, including fast motion and small objects, are considered to enhance the challenge of the dataset.

For the evaluation metrics, we adopt the widely used \textbf{Success Rate (SR)}, \textbf{Precision Rate (PR)}, and \textbf{Normalized Precision Rate (NPR)}. Considering that efficiency is also critical for practical tracking systems, we use \textbf{Frames Per Second (FPS)} to assess the tracking speed.

\subsection{Implementation Details}  
The training of our Slow-Fast tracker can be divided into two stages. We first train our slow tracker and fast tracker for 50 epochs independently. The learning rate of the Flash ViT and GCN is 0.0004 and 0.0005, respectively. The weight decay is 0.0001, and the batch size is 38. 
Then, the joint fine-tuning strategy is adopted during the second stage to effectively integrate the two trackers into a unified system. We fine-tuned the unified tracker for 20 epochs. The learning rates for Flash ViT and GCN are set to 0.00004 and 0.0005, respectively.
The AdamW~\cite{loshchilov2018adamw} is selected as the optimizer in both stages of training. The weighting coefficients $\lambda_i$ ($i \in [1,4]$) of the loss functions are set to 1, 14, 1, and 0.1, respectively.
Our code is implemented in Python based on the PyTorch~\cite{paszke2019pytorch} framework. All experiments are conducted on a server equipped with an AMD EPYC 7542 32-Core CPU and an NVIDIA RTX 4090 GPU. More details can be found in our source code.

\begin{figure*}[!htp]
\centering
\includegraphics[width=\textwidth]{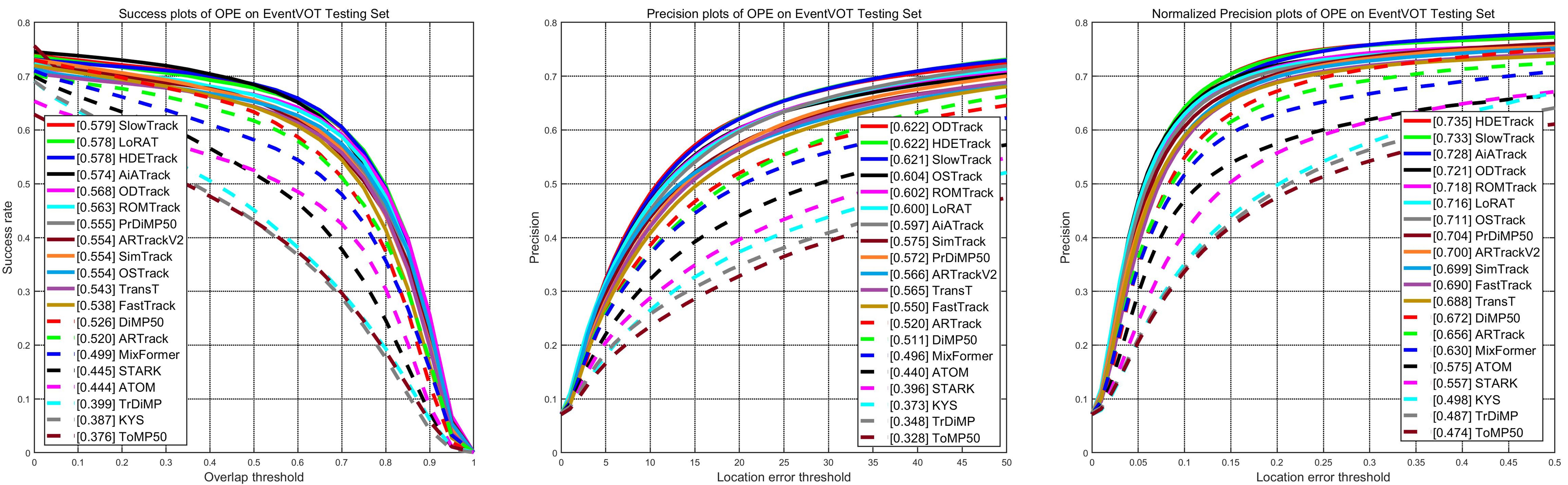}
\caption{Visualization of tracking results on the EventVOT dataset.}  
\label{PRSRNPRfig}
\end{figure*}

\subsection{Comparison on Public Benchmarks}

\noindent $\bullet$ \textbf{Results on FE240hz Dataset.~} 
As presented in Table~\ref{FE240table}, the baseline OSTrack achieves 57.1 and 89.3 on the SR and PR metrics, respectively. Encouragingly, our fast tracker, which contains only 6-layer ViTs, already exceeds the baseline performance to a certain extent, achieving 59.1 on SR and 91.7 on PR. Furthermore, our slow tracker establishes a new SOTA by surpassing strong competitors such as HDETrack and ATOM, attaining SR and PR scores of 59.9 and 92.7, respectively. These results comprehensively validate the effectiveness of the proposed Slow-Fast tracking framework in advancing event-based visual object tracking.

\begin{table}
\center
\small     
\caption{Experimental results (SR/PR) on FE240hz dataset.} 
\label{FE240table}
\resizebox{\columnwidth}{!}{
\begin{tabular}{cccccccccc}
\hline 
\textbf{STNet~\cite{zhang2022spiking}}  &\textbf{TransT~\cite{chen2021transt}}  &\textbf{STARK~\cite{yan2021Stark}}    &\textbf{PrDiMP~\cite{martin2020PrDimp}}  &\textbf{EFE~\cite{zhang2021fe108}}    &\textbf{SiamFC++~\cite{xu2020siamfc++}}    \\ 
58.5/89.6      &56.7/89.0        &55.4/83.7       &55.2/86.8        &55.0/83.5       &54.5/85.3 \\ 
\hline 
\textbf{DiMP~\cite{goutam2019Dimp}}   &\textbf{ATOM~\cite{danelljan2019atom}}  &\textbf{OSTrack~\cite{ye2022Ostrack}} &\textbf{HDETrack~\cite{wang2024event}}  &\textbf{Ours-Fast} &\textbf{Ours-Slow}     \\ 
53.4/88.2      &52.8/80.0         & 57.1/89.3        &59.8/92.2          &59.1/91.7    &59.9/92.7\\         
\hline 
\end{tabular}
}
\end{table}

\begin{table}
\center
\small   
\caption{Overall tracking performance on COESOT dataset. } 
\label{COESOT_results}
\begin{tabular}{l|l|c|cccc}
\hline 
\textbf{No.} & \textbf{Trackers} & \textbf{Source}   & \textbf{SR}  &\textbf{PR}   &\textbf{NPR} \\
\hline
01 &\textbf{ TrDiMP~\cite{wang2021TrDiMP} }   & CVPR21     &50.7       &59.2      &58.4           \\ 
02 &\textbf{ ToMP50~\cite{mayer2022Tomp}  }   &  CVPR22   &46.3       &55.2      &56.0           \\ 
03 &\textbf{ OSTrack~\cite{ye2022Ostrack} }   &  ECCV22   &50.9       &61.8       &61.5          \\ 
04 &\textbf{ AiATrack~\cite{gao2022AIa}   }   &ECCV22   &50.6       &59.5       &59.2           \\ 
05 &\textbf{ STARK~\cite{yan2021Stark}   }   &  ICCV21    &40.8      &44.5      &46.1           \\ 
06 &\textbf{ TransT~\cite{chen2021transt}   }   &  CVPR21     &45.6       &54.3       &54.2           \\ 
07 &\textbf{ DiMP50~\cite{goutam2019Dimp}   }  &  ICCV19     &53.8      &64.8       &65.1           \\ 
08 &\textbf{ PrDiMP~\cite{martin2020PrDimp}   }  &  CVPR20     &47.5       &57.8      &57.9           \\ 
09 &\textbf{ KYS~\cite{bhat2022SKys}   }   &   ECCV20      &42.6       &52.7       &52.1           \\ 
10 &\textbf{ MixFormer~\cite{cui2022mixformer}   }   & CVPR22   &44.4     &50.2      &51.1           \\ 
11 &\textbf{ ATOM~\cite{danelljan2019atom}   }   & CVPR19    &42.1      &50.4        &51.3          \\ 
12 &\textbf{ SimTrack~\cite{chen2022SimTrack}   }   & ECCV22  & 48.3      &55.7       &56.6           \\  
13 &\textbf{ HDETrack~\cite{wang2024event}  }    &CVPR24            &53.1       &64.1      &64.5          \\ 
\hline
14 &\textbf{Ours-Fast} & - &49.3   &59.1   &59.8 \\
15 &\textbf{Ours-Slow} & - &51.8   &62.9   &62.9 \\
\hline
\end{tabular}
\end{table}

\noindent $\bullet$ \textbf{Results on COESOT Dataset.~}
As shown in Table~\ref{COESOT_results}, we also report our tracking results on the large-scale RGB-Event tracking dataset COESOT. To ensure a fair comparison, only event modality data is used as input to the network. Experimental results demonstrate that our fast tracker surpasses conventional CNN-based tracking methods, such as ATOM and PrDiMP, achieving 49.3 and 59.1 on SR and PR, respectively. Furthermore, our Slow Tracker exhibits notable improvements over OSTrack, with gains of 0.9 and 1.1 in SR and PR, respectively. Additionally, when compared with other SOTA trackers, like AiATrack and ODTrack, our approach maintains a clear performance advantage. These experimental results fully validate the effectiveness of our Slow-Fast tracker in event-based visual object tracking tasks.


\noindent $\bullet$ \textbf{Results on EventVOT Dataset.~}
As shown in Table~\ref{EventVOTtable} and Fig.~\ref{PRSRNPRfig}, we conduct a comprehensive comparison with advanced tracking algorithms on the EventVOT dataset. Our fast tracker achieves comparable accuracy while significantly outperforming previous methods in inference speed (126 FPS). Meanwhile, by performing multi-scale fusion, we significantly enhance the performance of the slow tracker. Compared to methods such as ROMTrack and LoRAT, our slow tracker shows a substantial advantage in accuracy, achieving 57.9, 62.1, and 73.3 on SR, PR, and NPR, respectively. Overall, these results demonstrate the strong performance of our approach under both resource-constrained and resource-rich conditions, making it adaptable to different real-world applications.

\begin{table}
\centering
\small   
\caption{Overall Tracking Performance on EventVOT Dataset. } 
\label{EventVOTtable}
\resizebox{\columnwidth}{!}{ 
\begin{tabular}{l|l|c|lll|ll}
\hline \toprule [0.5 pt]
\textbf{No.} & \textbf{Trackers} & \textbf{Source}   & \textbf{SR}  &\textbf{PR}   &\textbf{NPR}  &\textbf{Params}  &\textbf{FPS}\\
\hline
01    &  \textbf{ DiMP50~\cite{goutam2019Dimp} }  &  ICCV19       &\ 52.6   &\ 51.1   &\ 67.2   &\ 26.1  &\ 43  \\
02    &  \textbf{ ATOM~\cite{martin2019Atom}   }   & CVPR19     &\ 44.4   &\ 44.0   &\ 57.5   &\ 8.4   &\ 30  \\
03    &  \textbf{ PrDiMP~\cite{martin2020PrDimp}}  &  CVPR20       &\ 55.5   &\ 57.2   &\ 70.4   &\ 26.1   &\ 30  \\
04    &  \textbf{ KYS~\cite{bhat2022SKys}   }   &   ECCV20         &\ 38.7   &\ 37.3   &\ 49.8   &\ --   &\ 20  \\ 
05    &  \textbf{ TrDiMP~\cite{wang2021TrDiMP} } & CVPR21     &\ 39.9   &\ 34.8   &\ 48.7  &\ 26.3   &\ 26   \\ 
06    &  \textbf{ STARK~\cite{yan2021Stark}   }   &  ICCV21     &\ 44.5   &\ 39.6  &\ 55.7   &\ 28.1   &\ 42  \\ 
07    &  \textbf{ TransT~\cite{chen2021transt}   }   &  CVPR21     &\ 54.3  &\ 56.5  &\ 68.8   &\ 18.5   &\ 50  \\ 
08    &  \textbf{ ToMP50~\cite{mayer2022Tomp}  }   &  CVPR22   &\ 37.6   &\ 32.8   &\ 47.4   &\ 26.1   &\ 25  \\ 
09    &  \textbf{ OSTrack~\cite{ye2022Ostrack} }   &  ECCV22   &\ 55.4  &\ 60.4   &\ 71.1   &\ 92.1   &\ 105  \\
10    &  \textbf{ AiATrack~\cite{gao2022AIa}   }   &  ECCV22     &\ 57.4   &\ 59.7   &\ 72.8   &\ 15.8   &\ 38  \\ 
11    &  \textbf{ MixFormer~\cite{cui2022mixformer}   }   & CVPR22     &\ 49.9   &\ 49.6   &\ 63.0   &\ 35.6   &\ 25  \\
12    &  \textbf{ SimTrack~\cite{chen2022SimTrack}   }   & ECCV22     &\ 55.4   &\ 57.5  &\ 69.9   &\ 57.8   &\ 40  \\ 
13    &  \textbf{ ROMTrack~\cite{cai2023robust} } &ICCV23  &\ 56.3  &\ 60.2 &\ 71.8 &\ 92.1   &\ 62 \\
14    &  \textbf{ ARTrack~\cite{wei2023autoregressive} } &CVPR23  &\ 52.0  &\ 52.0 &\ 65.6 &\ 172.0   &\ 26  \\ 
15    &  \textbf{ ODTrack~\cite{zheng2024odtrack}}      &AAAI24  &\ 56.8   &\ 62.2  &\ 72.1   &\ 92.0   &\ 32  \\ 
16    &  \textbf{ ARTrackV2~\cite{bai2024artrackv2} } &CVPR24   &\ 55.4 &\ 56.6 &\ 70.0  &\ 101.0    &\ 94  \\
17    &  \textbf{ LoRAT~\cite{lin2024tracking} } &ECCV24  &\ 57.8 &\ 60.0 &\ 71.6 &\ 99.0    &\ 209  \\
18    &  \textbf{ HDETrack~\cite{wang2024event}}    &CVPR24 &\ 57.8   &\ 62.2  &\ 73.5   &\ 92.1 &\ 105 \\ 
\hline
19    &  \textbf{ Ours-Fast }      &- &\ 53.8   &\ 55.0 &\ 69.0  &\ 50.4   &\ 126   \\ 
20    &  \textbf{ Ours-Slow }      &-  &\ 57.9  &\ 62.1  &\ 73.3 \  &\ 93.4  &\ 70   \\ 
\hline \toprule [0.5 pt]
\end{tabular}
}
\end{table}

\subsection{Ablation Study} 

\noindent $\bullet$ \textbf{Analysis of Input Data.~}
In this section, we present a comprehensive and independent ablation study of our experimental results. As shown in Table~\ref{CAResults}, we begin by investigating the impact of multi-view fusion on the final performance, specifically evaluating the effectiveness and necessity of incorporating event point data. Compared to using only event frames, integrating high-temporal-resolution event stream data consistently improves performance for both the slow and fast trackers on the EventVOT dataset. Furthermore, applying knowledge distillation during the fine-tuning of the fast tracker leads to additional accuracy gains, achieving competitive results (SR: 53.8, PR: 55.0, NPR: 69.0) that approach the baseline OSTrack while maintaining low latency. These findings demonstrate the effectiveness of our training strategy and the proposed Slow-Fast tracking framework.

\begin{table}
\center
\small     
\caption{Component Analysis results (SR, PR, NPR) on the EventVOT dataset.} 
\label{CAResults} 
\resizebox{\columnwidth}{!}{
\begin{tabular}{c|ccc|ccc} 		
\hline 
\textbf{Fast/Slow} & \textbf{Event frames} &\textbf{Event points}  &\textbf{Fine-tuning}    &\textbf{SR}   &\textbf{PR}   & \textbf{NPR}\\
\hline 
\multirow{3}{*}{Fast} & \cmark &        &              & 52.5  & 53.8  & 67.1      \\
                      & \cmark & \cmark &              & 52.8  & 54.2  & 68.1      \\
                     & \cmark &  \cmark      & \cmark       & 53.8  & 55.0  & 69.0      \\
\hline
\multirow{2}{*}{Slow} & \cmark &        &              & 55.4  & 60.4  & 71.1      \\
                    & \cmark & \cmark &        & \textbf{57.9} & \textbf{62.1} & \textbf{73.3} \\
\hline
\end{tabular}
}
\end{table}

\begin{table}
\centering
\small
    \caption{Ablation studies on Dimensions of GCN Feature on EventVOT dataset.} 
    \label{eventvot_ablation_gcn}
    \begin{tabular}{c|c|lll} 
    \hline 
    \textbf{\# GCN feature dim}  &\textbf{Slow/Fast}   &\textbf{SR}   & \textbf{PR}  & \textbf{NPR}  \\
    \hline
    \text{[1, 8, 16]}  &\multirow{5}{*}{Slow}     &56.8   &61.0   &72.0  \\
    \text{[1, 16, 32]}                             & &57.0   &61.0    &72.4   \\
    \textbf{\text{[1, 16, 64]}}               &    &\textbf{57.9}   &\textbf{62.1}   &\textbf{73.3}  \\
    \text{[1, 16, 128]}                             &    &56.5    &60.4  &71.8  \\
    \text{[1, 32, 128]}                             &   &56.2    &59.3  &71.1  \\
    \hline
    \end{tabular}
\end{table}

\begin{figure}
\centering
\includegraphics[width=0.48\textwidth]{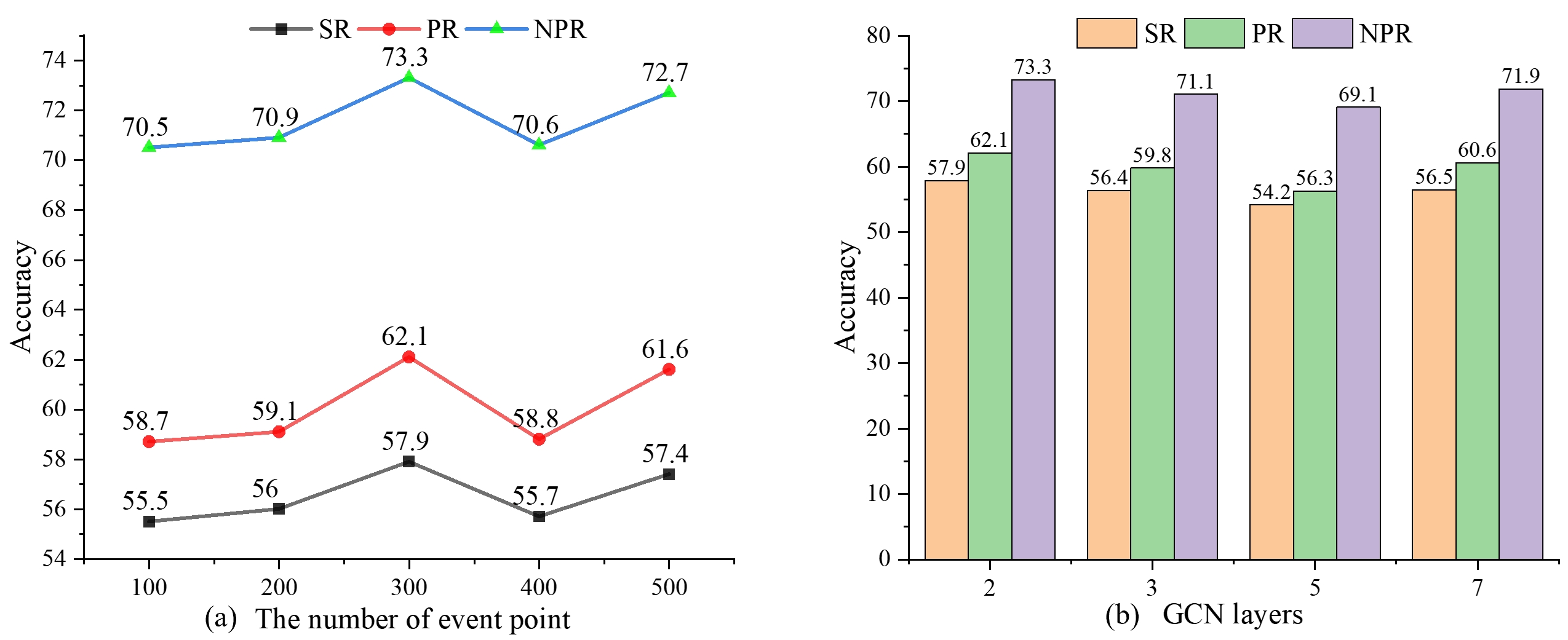}
\caption{(a). Analysis of the number of event points. (b). Analysis of the number of GCN layers.}
\label{ablation_study}
\end{figure}

\noindent $\bullet$ \textbf{Analysis of the Number of Event Points.~}
As illustrated in Fig.~\ref{ablation_study} (a), we conducted an analysis on the number of event points in the input event streams (i.e., the number of event points retained per frame after downsampling). Our finding reveals that when the number of event points is insufficient (e.g., retaining only 100 event points), the information representation capability becomes severely limited, resulting in suboptimal performance. Conversely, retaining excessive event points may introduce substantial noise that adversely affects the experimental results. Therefore, in this work, we select 300 event points as the input for the event streams to achieve optimal performance.

\noindent $\bullet$ \textbf{Analysis of the Number of GCN Layers.~}
As shown in Fig.~\ref{ablation_study} (b), we also performed a comprehensive analysis of the number of GCN layers to determine the optimal network depth for modeling event points-based graph structure representations. The result demonstrates that stacking too many GCN layers (e.g., 3, 5, or 7 layers) not only increases computational overhead but also leads to the over-smoothing problem, ultimately degrading overall performance. Therefore, we employ a 2-layer GCN architecture for extracting graph features from event points, which achieves more effective feature modeling while maintaining lower network complexity.

\noindent $\bullet$ \textbf{Analysis on Dimensions of GCN Feature.~}
As presented in Table~\ref{eventvot_ablation_gcn}, we further investigate the impact of varying hidden layer dimensions in the GCN architecture. The experimental results show that when employing a 2-layer GCN to process event point features, the best performance is achieved with a progressive dimensional configuration of 1→16→64. This configuration effectively balances feature representation capacity and computational efficiency. In comparison, smaller feature dimensions lead to inadequate feature encoding and information loss, whereas excessively large dimensions increase the model’s susceptibility to noise.

\begin{figure}[!htp]
\includegraphics[width=\linewidth]{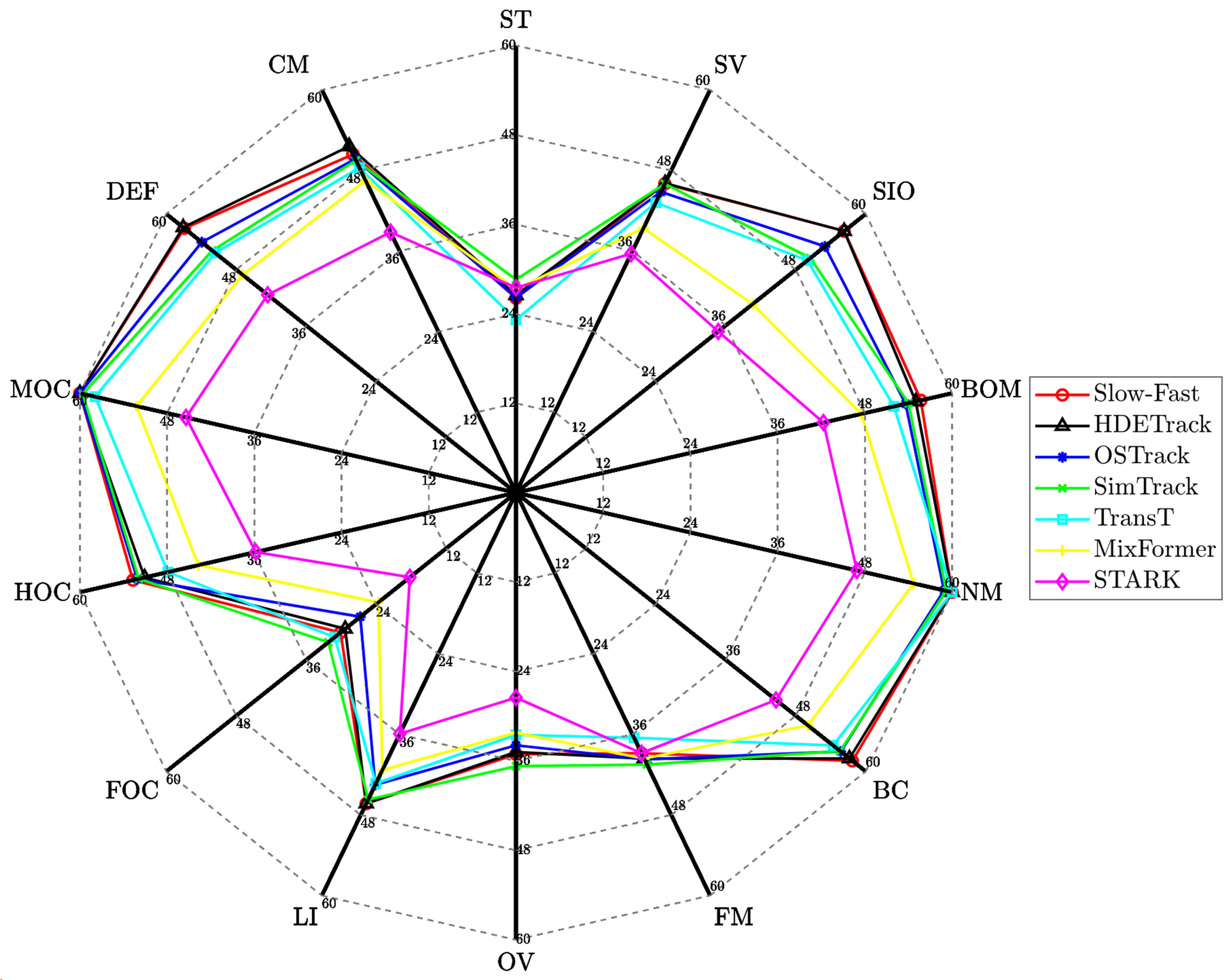}
\caption{Tracking results (SR) under each challenging factor.} 
\label{attributeResults}
\end{figure}

\noindent $\bullet$ \textbf{Analysis of Fusion Layer.~}
In the fast tracker, we employ a simple yet effective cross-view fusion method to integrate multi-view features. However, applying the same approach to the slow tracker yields only marginal improvements (as shown in the 12th-layer Transformer results in Table~\ref{eventvot_ablation_fusion}), falling short of our design expectations. To address this limitation, we propose a multi-scale fusion strategy that combines event point and event frame information to improve the robustness of the slow tracker. Experimental results confirm that this multi-scale fusion approach significantly enhances the tracking performance of the slow tracker.

\begin{table}
\center
    \caption{Ablation studies of Fusion Layer.} 
    \label{eventvot_ablation_fusion}
    \begin{tabular}{l|c|ll} 
    \hline 
    \textbf{\# Fusion layer}  &\textbf{Slow/Fast}   &\textbf{SR}   & \textbf{PR}   \\
    \hline
    \text{0$^{th}$-layer Former}    &\multirow{3}{*}{Slow}   &55.9   &60.8     \\
    \text{12$^{th}$-layer Former}    &    &55.8    &60.6   \\
    \textbf{\text{Multi-scale fusion}}   &    &\textbf{57.9}   &\textbf{62.1}     \\
    \hline
    \end{tabular}
\end{table}

\begin{table}
\center
    \caption{Ablation studies of Graph Construction.} 
    \label{Graph_construction}
    \begin{tabular}{l|c|ll} 
    \hline 
    \textbf{\# Graph construction}  &\textbf{Slow/Fast}   &\textbf{SR}   & \textbf{PR}  \\
    \hline
    \text{Random}    &\multirow{3}{*}{Slow}   &55.9   &60.8    \\
    \text{Radius graph}    &    &57.3    & 61.3  \\
    \textbf{\text{K-NN}}   &    &\textbf{57.9}   &\textbf{62.1}    \\
    \hline
    \end{tabular}
\end{table}

\noindent $\bullet$ \textbf{Analysis of Graph Construction.~}
The method of graph construction is a crucial factor for event-based graph data. As shown in the Table.~\ref{Graph_construction}, we investigate three different graph construction approaches here to find a suitable way to represent event points. It can be observed that the random graph construction method offers limited benefits. Both the neighborhood radius-based graph construction and the K-nearest neighbors (K-NN)-based graph construction methods effectively improve tracking performance. Here, we adopt the K-NN-based graph construction approach due to its stronger graph representation capability and high construction efficiency.

\begin{figure*}
\center
\includegraphics[width=\textwidth]{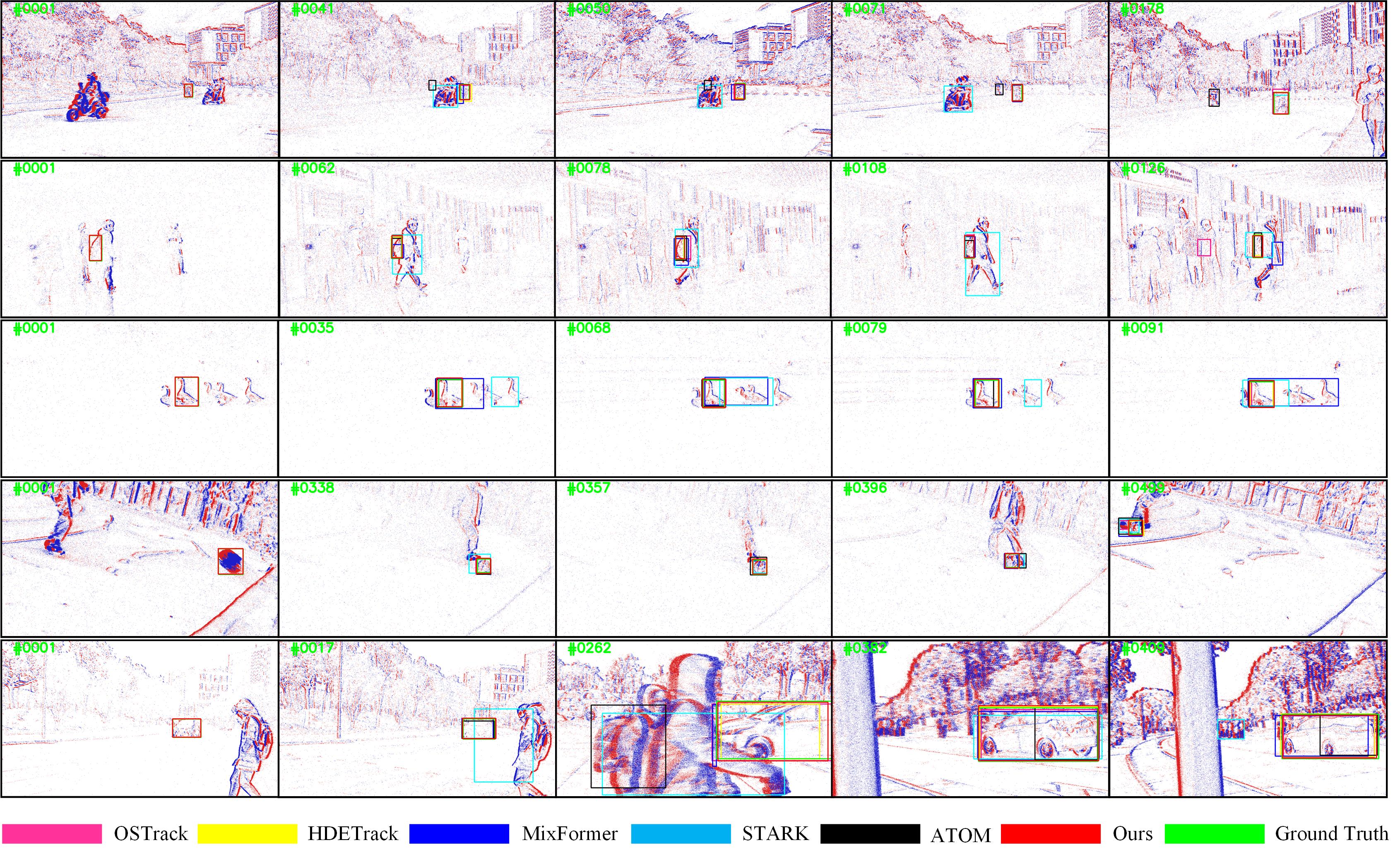}
\caption{Visualization of the tracking results of our slow tracker and other SOTA trackers.}  
\label{trackingResults}
\end{figure*}

\begin{figure}
\center
\includegraphics[width=3in]{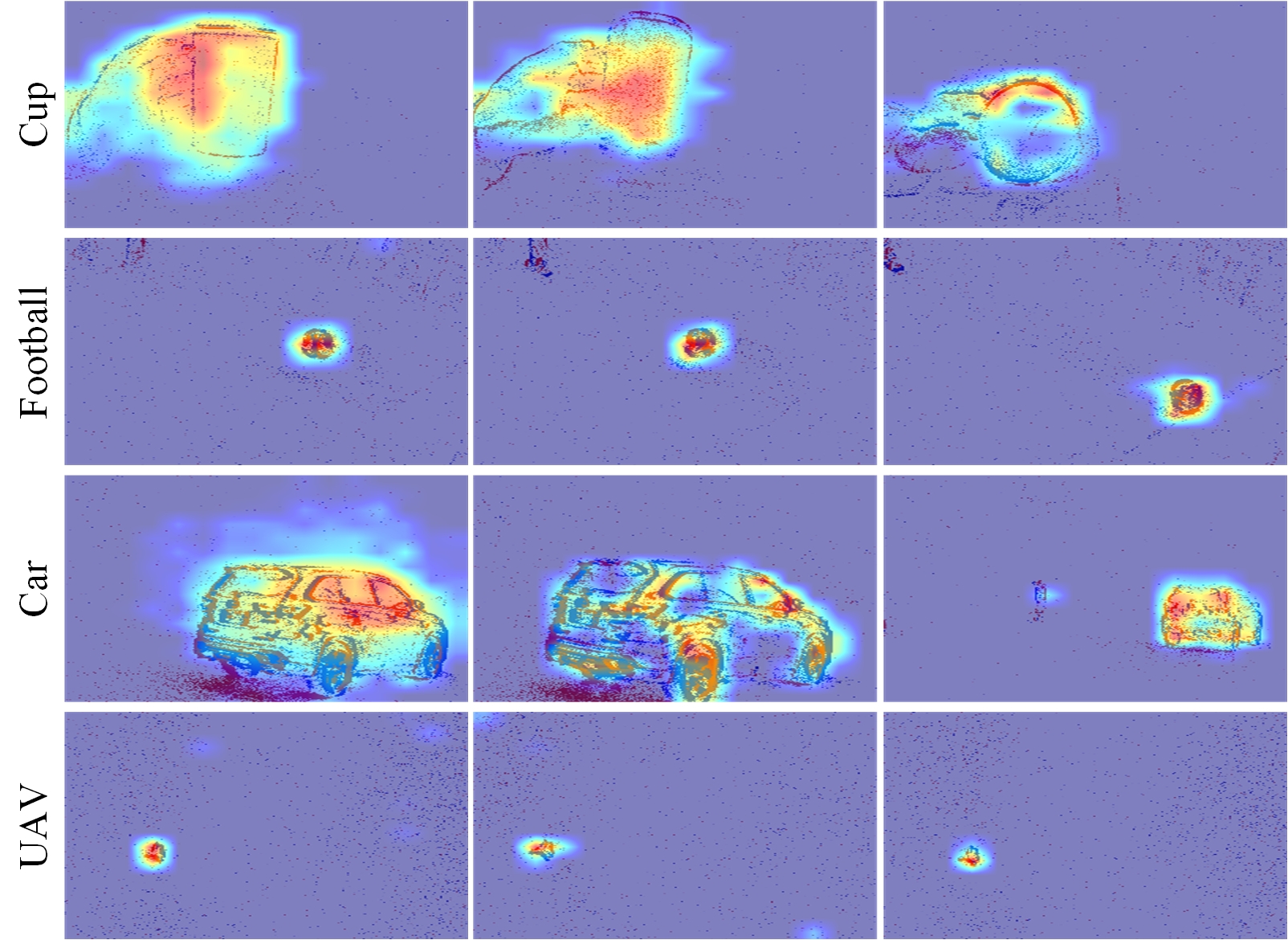}
\caption{Attention maps predicted by our slow tracker.}  
\label{attention_map}
\end{figure}

\noindent $\bullet$ \textbf{Results Under Each Attribute.~}
As shown in Fig.~\ref{attributeResults}, we also report the results of our slow tracker and other state-of-the-art trackers under each challenging scenario. We can see that our slow tracker achieves superior performance under several challenging attributes, including background clutter (BC), similar object motion (BOM), and heavily occluded (HOC), outperforming algorithms like HDETrack. Furthermore, it maintains competitive performance across other challenge categories, surpassing most mainstream tracking methods. These findings validate the strong robustness of our slow tracker in complex real-world scenarios.

\subsection{Parameter Analysis} 
As detailed in Table~\ref{EventVOTtable}, we also perform a comprehensive parameter analysis of our proposed Slow-Fast tracking architecture to provide a more intuitive understanding of our model. The analysis reveals that our slow tracker achieves robust tracking performance with 93.4M parameters when sufficient computational resources are available. In contrast, the fast tracker demonstrates remarkable efficiency, achieving an inference speed of 126 FPS with merely 50.4M parameters, making it particularly suitable for resource-constrained environments. Furthermore, due to the high-temporal-resolution event stream data, our fast tracker can rapidly output multiple tracking results (i.e., output three results per frame with 56 FPS, each tracking result is generated in approximately \textbf{6 ms}), fulfilling the critical requirements of latency-sensitive applications. Therefore, we believe that the Slow-Fast architecture provides flexible deployment options across different operational scenarios.

\subsection{Visualization} 
\noindent $\bullet$ \textbf{Tracking Results.}
To gain a more comprehensive understanding of our tracking algorithm, we perform a visual comparison between our proposed slow tracker and other SOTA tracking methods. As illustrated in Fig.~\ref{trackingResults}, we present qualitative tracking results of our slow tracker alongside those of OSTrack, MixFormer, and STARK. It can be observed that our method exhibits superior stability and more robust performance in real-world tracking scenarios compared to these SOTA approaches.

\begin{figure}
\center
\includegraphics[width=3in]{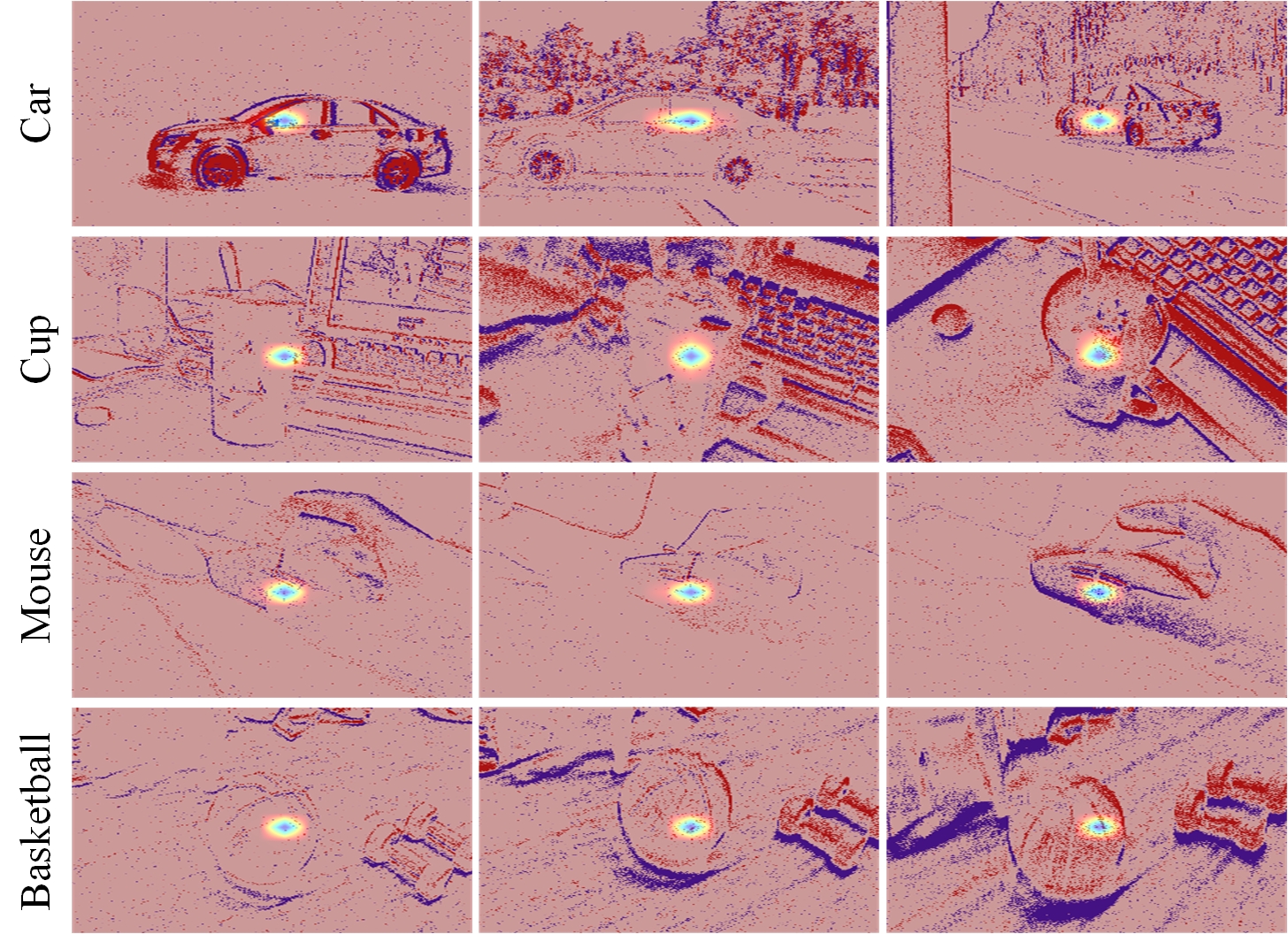}
\caption{Response maps predicted by our slow tracker.}  
\label{responseMaps}
\end{figure}

\noindent $\bullet$ \textbf{Attention Heat Maps and Response Maps.}
As illustrated in Fig.~\ref{attention_map}, we visualize the attention heatmaps generated by our slow tracker. In these maps, the intensity of the red regions reflects the magnitude of attention weights, with deeper hues indicating stronger focus. Notably, our model consistently attends to the target object throughout the tracking process, effectively highlighting discriminative features while suppressing background noise.
Fig.~\ref{responseMaps} further presents the response maps produced by the model, showcasing its localization capability. The highlighted areas correspond to peak response values, indicating the predicted target positions. As shown, our method achieves precise and stable target localization, characterized by sharp and well-defined response peaks.
These visualizations provide strong qualitative evidence of the robustness and effectiveness of our slow tracking approach.

\subsection{Limitation Analysis}  
Although our Slow-Fast tracking framework demonstrates robust adaptability across different real-world scenarios, there also remains room for improvement.
Firstly, the current model is not explicitly designed to address various challenging factors, which limits its optimal performance in difficult tracking scenarios (e.g., sparse events caused by slow-moving objects). We propose to leverage large multimodal models to analyze the challenge-specific attributes embedded in video sequences and implement corresponding adaptive strategies.
Secondly, the model's performance still heavily relies on training data distribution, leading to degraded performance in open-world scenarios beyond the coverage of training data. To mitigate this, we will employ style transfer techniques to enhance the model's generalization capability in open-world environments.

\section{Conclusion}  
In this paper, we propose a novel Slow-Fast tracking framework for event-based visual object tracking. The framework comprises a slow tracker tailored for high-precision performance and a fast tracker optimized for low-latency tracking. By effectively incorporating event point-based graph-structured data into both trackers, our approach enhances temporal perception and robustness. Additionally, a simple yet effective supervised fine-tuning strategy seamlessly integrates the two trackers and further enhances the performance of the fast tracker through a knowledge distillation strategy. The proposed framework demonstrates reliable tracking in resource-rich environments while maintaining millisecond-level low latency in resource-constrained scenarios.

{
    \small
    \bibliographystyle{ieeenat_fullname}
    \bibliography{bib}
}


\end{document}